\pdfoutput=1
\documentclass[11pt]{article}

\usepackage[preprint]{acl}

\usepackage{times}
\usepackage{latexsym}

\usepackage[T1]{fontenc}
\usepackage[utf8]{inputenc}

\usepackage{microtype}

\usepackage{inconsolata}
\usepackage{graphicx}
\usepackage{amsmath}
\usepackage{booktabs}

\title{RevOrder: A Novel Method for Enhanced Arithmetic in Language Models}

\author{Si Shen \\
  Nanjing University of Science \\ and Technology,  \\ Nanjing, China \\\And
  Peijun Shen \\
  Henan University, \\ Kaifeng, China \\
  \texttt{shenpeijun@henu.edu.cn} \\
  \\\And
  Danhao Zhu \\
  Jiangsu Police Institute, \\ Nanjing, China \\
  \texttt{229369897@qq.com} \\} 

\begin{document}
\maketitle
\begin{abstract}
This paper presents RevOrder, a novel technique aimed at improving arithmetic operations in large language models (LLMs) by reversing the output digits in addition, subtraction, and n-digit by 1-digit (nD by 1D) multiplication tasks. Our method significantly reduces the Count of Sequential Intermediate Digits (CSID) to $\mathcal{O}(1)$, a new metric we introduce to assess equation complexity. Through comprehensive testing, RevOrder not only achieves perfect accuracy in basic arithmetic operations but also substantially boosts LLM performance in division tasks, particularly with large numbers where traditional models struggle. Implementation of RevOrder is cost-effective for both training and inference phases. Moreover, applying RevOrder to fine-tune the LLaMA2-7B model on the GSM8K math task results in a considerable improvement, reducing equation calculation errors by 46\% and increasing overall scores from 41.6 to 44.4. \footnote{Corresponding authors: Danhao Zhu} \footnote{The data and code for this paper are available on Github.}
\end{abstract}

\section{Introduction}
Large language models (LLMs) have gained significant attention in recent years, excelling in natural language understanding and generation tasks~\cite{zhao2023survey}. Despite their advancements, the leading models like ChatGPT~\cite{openai2022chatgpt} and GPT-4~\cite{openai2023gpt4} struggle with basic arithmetic, particularly with large digits. The GPT-4 website service\footnote{https://chat.openai.com/, 2024-1-26} addresses this by switching to external Python tools, as depicted in Fig. 1(a). This shift not only adds a cumbersome step but also leads to excessive token usage, significantly disrupting the language processing flow and efficiency.

\begin{figure}
    \centering
    \includegraphics[width=0.5\textwidth]{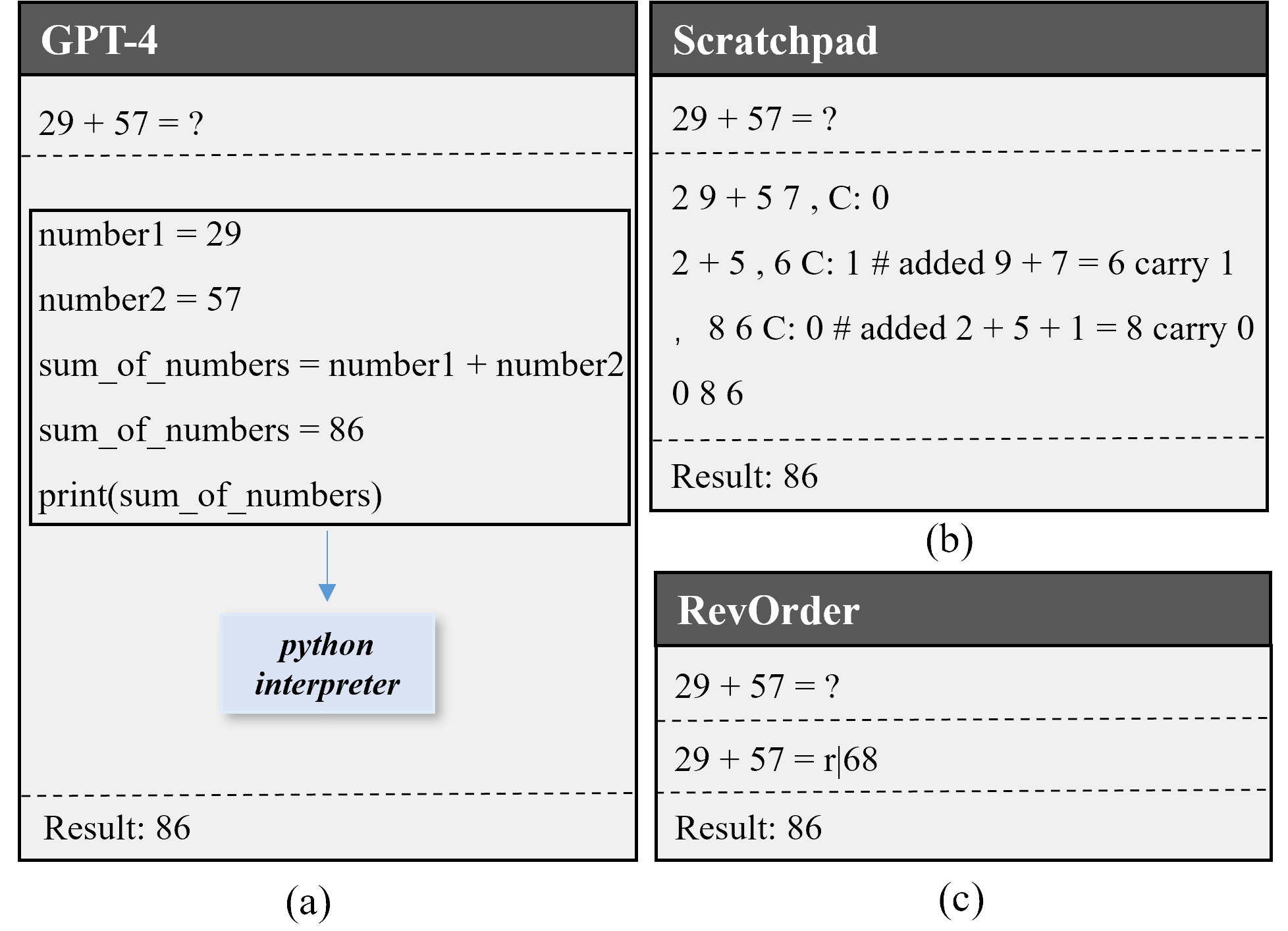}  % Adjust the width as needed
    \caption{An illustration of performing addition using various methods. In the RevOrder method, the 'r|' symbol indicates that the subsequent digits are presented in reverse order.}
    \label{fig:my_label}
\end{figure}

Arithmetic reasoning has long focused on solving arithmetic problems with LMs~\cite{lu2022survey}. Typically, LMs generate the solutions step-by-step in a chain-of-thought (COT) manner, common in reasoning tasks~\cite{wei2022chain,kojima2022large,zhou2022least}. For instance, ~\citet{nye2021show} used a 'Scratchpad' to generate intermediate steps, achieving high accuracy in 8D addition tasks, as shown in Fig. 1(b). Similar methods are applied for subtraction, multiplication, division, and other arithmetic operations~\cite{liu2023goat,yang2023gpt}.

However, practical application of arithmetic reasoning in LMs faces significant challenges. Firstly, LMs lack consistency in providing accurate results, and there is no established theory to measure equation complexity or to determine if an equation is within an LM's capabilities. For example, ~\citet{liu2023goat} posited that addition is learnable by LLMs, but their experiments with large-digit addition contained minor errors. Secondly, current decomposition methods are token-intensive, making them more expensive than tool-based solutions during inference. Even for a simple 2D addition, the Scratchpad method~\cite{nye2021show}, shown in Fig. 1(b), is not more token-efficient than Python tools (Fig. 1(a)).

To address these challenges, we introduce two novel concepts. First, we propose the Count of Sequential Digits (CSID) as an indicator to measure the difficulty of arithmetic equations. A larger CSID suggests more omitted reasoning steps, indicating a more complex equation. We demonstrate that the CSID complexity grows at $\mathcal O(n)$ for addition and subtraction, where $n$ is the digit count. Empirical evidence suggests that advanced language models struggle considerably with high-CSID problems. This indicates a notable limitation: LLMs are unreliable in directly producing results for even basic arithmetic tasks, such as single additions or subtractions, when the digits involved are large.

Second, we propose RevOrder, a technique that reduces the CSID to a constant 1 for addition, subtraction, and nD by 1D multiplication operations. Illustrated in Fig. 1(c), RevOrder reverses the output order of addition. This approach aligns with the natural human reasoning sequence, where higher-order digits are resolved after the lower ones. Unlike previous methods such as Scratchpad~\cite{nye2021show}, RevOrder requires virtually no additional tokens for these basic operations. Building upon these, we can construct more complex operations with significantly reduced token usage.

RevOrder, evaluated on the Big-bench arithmetic task~\cite{srivastava2022beyond} and an expanded set with larger digits, achieved 100\% accuracy in addition, subtraction, multiplication, and low-digit division tasks, and nearly 100\% in large-digit division, outperforming baseline methods. The experimental section highlights its training and inference efficiency. Finetuning LLAMA2~\cite{touvron2023llama} with RevOrder on the GSM8K dataset~\cite{cobbe2021training} significantly improved equation accuracy and overall scores (from 88.9\% to 94.1\%, and 41.6 to 44.4, respectively). These results affirm RevOrder's effectiveness and token economy in a range of arithmetic tasks, especially in addition and subtraction.

Section 2 reviews related work, Section 3 introduces the CSID metric, Section 4 details the RevOrder technique, Section 5 reports on experiments on arithmetic calculation, Section 6 discusses finetuning on GSM8K, and Section 7 concludes the paper.

\section{Related Works}
Arithmetic ability, a cornerstone of mathematics, has long served as a benchmark for assessing model capabilities, evolving from statistical methods~\cite{hosseini2014learning} through machine learning techniques~\cite{kushman2014learning}, deep learning approaches~\cite{wang2017deep} to LLM methods~\cite{wei2022emergent}.

While scaling laws for LLMs suggest that model capacity increases with model size, compute resources, and training data~\cite{kaplan2020scaling,hoffmann2022training}, LLMs often struggle to directly generate arithmetic results. Consequently, step-by-step arithmetic reasoning methods have been developed. ScratchPad~\cite{nye2021show} introduces this concept for additions, achieving near-perfect accuracy on 8D addition tasks. This idea has since been expanded to more complex operations, such as multiplication and division~\cite{liu2023goat,yang2023gpt}. These complex operations depend on the assumption that LLMs can efficiently perform basic operations such as addition and subtraction. Otherwise, token usage quickly becomes unsustainable. However, these so-called basic operations often fail to achieve 100\% accuracy with large digits, making the more complex operations built upon them even more prone to error. Our CSID theory provides a framework to assess the complexity of equations, showing that LLMs' ability to perform basic operations diminishes as digit size grows. Conversely, RevOrder introduces an efficient method to keep equations' CSID low, ensuring their manageability within constrained token budgets.

Given the limitations and high token consumption of previous arithmetic reasoning methods, more pragmatic solutions have emerged, such as utilizing external tools or programming~\cite{schick2023toolformer,chen2022program,gao2023pal}. RevOrder stands out by offering reliability and efficiency in addition and subtraction, positioning itself as a resource-saving alternative to these methods.

\section{Sequential Intermediate Digits in Arithmetic Computation}
Arithmetic reasoning in language models (LMs) is challenging, mainly due to the sequential prediction of digits. This complexity is exacerbated when contextual digits required for accurate predictions are not inferred from previous steps. For example, in addition, LMs may predict higher-order digits before lower-order ones, contradicting the logical computation order. This paper introduces a novel metric to quantify and understand this complexity.

\subsection{Definition of Sequential Intermediate Digits (SIDs)}
A \textit{Sequential Intermediate Digit} (SID) is a numeral crucial for the accurate prediction of the next digit in a sequence, yet not present in the preceding sequence. Within the framework of chain-of-thought reasoning, SIDs represent indispensable steps that, despite being missing, are vital for the computational process. Consequently, the Count of SIDs (CSIDs) is employed as a metric to assess the complexity of a generation step, with a higher CSID denoting a more demanding and intricate task. The CSID of an equation is thus defined as the maximum CSID required for generating each step of the result.

The primary types of SIDs include:
\begin{itemize}
  \item Carry-over or borrow digits in addition and subtraction. For example, in $123 + 179 = 302$, the digit 3 in the hundreds place requires the carry-over from the tens and units places, resulting in a maximum CSID of 2.
  \item Digits from omitted reasoning steps, such as the intermediate sum \(3\) in \(1+2+4=7\).
\end{itemize}

It is postulated that basic operations like 1D by 1D addition, subtraction, multiplication, division, counting, and copying do not require SIDs, as their straightforward nature falls within the capabilities of modern LMs. Directly generating results for complex operations, such as multi-digit multiplication and division, requires more SIDs due to the omitted steps for decomposing these into multiple basic operations.

Reducing an equation's CSIDs, thereby lowering its solving difficulty, can be achieved by expanding the equation step-by-step in a chain-of-thought manner. For instance, the CSID for the calculation \(1+2+4=3+4=7\) is lower than for \(1+2+4=7\) because the intermediate sum \(3\) is included in the reasoning process, effectively reducing the number of SIDs.

\subsection{The CSIDs for Arithmetic Operations}
In our CSID analysis of standard arithmetic operations, which is akin to analyzing space or time complexity in algorithms, we focus on the worst-case scenario. Consider two numbers \(a = a_n a_{n-1} \ldots a_2 a_1\) and \(b = b_m b_{m-1} \ldots b_2 b_1\), resulting in \(c = c_t c_{t-1} \ldots c_2 c_1\), with \(m \le n\). When involving negative numbers, the minus sign '-' is also treated as a digit.

\begin{itemize}
  \item In addition and subtraction, the computation sequence \(a_n a_{n-1} \ldots a_2 a_1 \pm b_m b_{m-1} \ldots b_2 b_1 = c_t c_{t-1} \ldots c_2 c_1\) depends on each \(c_i\) involving \(a_i\), \(b_i\), and possibly \(c_{i-1}\) for carry-overs or borrows. Hence, the CSID for \(c_t\) includes all lower digits as SIDs, indicating a complexity of \(\mathcal{O}(n)\).
  \item For multiplication and division, the CSIDs are \(\mathcal{O}(n^2)\) and \(\mathcal{O}(n^2-m^2)\) respectively, as detailed in Appendix A.
\end{itemize}

\subsection{LLM Performance on Large CSID Equations}

\begin{figure}
    \centering
    \includegraphics[width=0.45\textwidth]{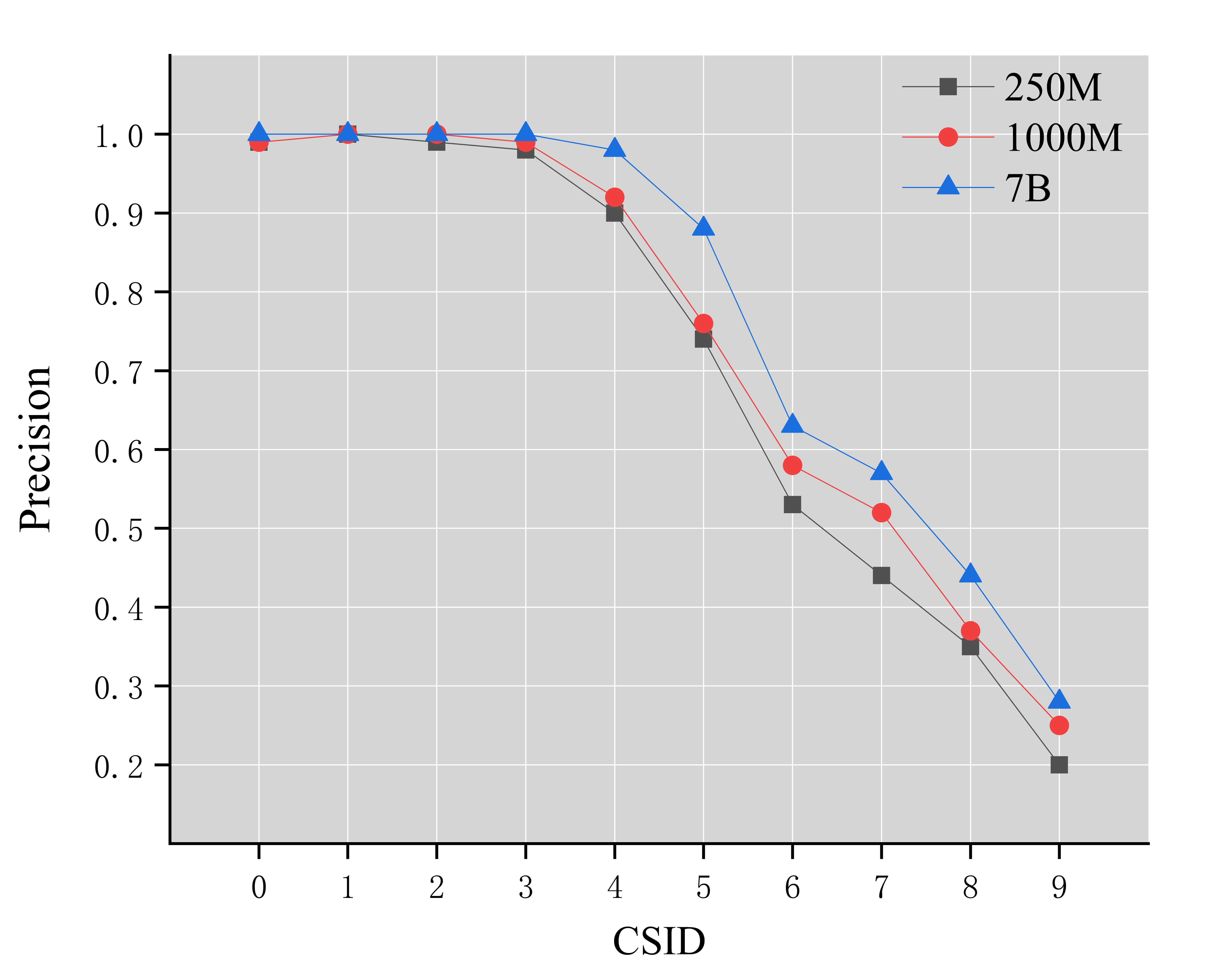}  % Adjust the width as needed
    \caption{LLM performance on equations with varying CSIDs.}
    \label{fig:my_label}
\end{figure}

We trained various models on arithmetic tasks involving 15D+15D calculations, maintaining identical hyper-parameters, training data, and training steps across all models to ensure a fair comparison. The test equations, strictly in 15D+15D format, were classified into various CSID levels according to the maximum number of continuous carry-over digits. The findings, as depicted in Fig. 2, demonstrate that:

\begin{itemize}
  \item CSID effectively measures the complexity of arithmetic equations, where the performance consistently declines with increasing CSIDs.
  \item Larger models exhibit improved performance on equations with higher CSIDs.
  \item The benefit of increasing model size diminishes on high CSID equations. For instance, a 7B model shows more significant improvement on equations with CSIDs of 4 and 5 than on those with 6-9. This trend suggests that even advanced LLMs, like GPT-4, encounter difficulties with large digit addition tasks. Given that CSIDs have a complexity of at least $\mathcal{O}(n)$, arithmetic problems quickly surpass the capacity of LLMs when dealing with large digits. Therefore, \textbf{LLMs cannot serve as reliable calculators for immediate result generation in complex arithmetic tasks}.
\end{itemize}

\section{RevOrder: Reversing the Order of Output Digits}
We introduce RevOrder, an innovative technique devised to maintain low CSID in equations, thereby ensuring their solvability by LMs. Additionally, RevOrder is designed to minimize token usage, enhancing overall efficiency.
\subsection{Addition and Subtraction}
For addition and subtraction, we reverse the output digits' order:
\begin{align*}
a \pm b &= r|c_1 c_2 \ldots c_t \\
&= c_t \ldots c_2 c_1
\end{align*}
Here, \( r| \) is a special token indicating that the followed digits are in a reversed order. To generate each \( c_i \) in \( r|c_1 c_2 \ldots c_t \), only \( a_i \), \( b_i \), and at most a SID for the carry-over or borrow number from \( c_{i-1} \) are required. Thus, both addition and subtraction only consume at most 1 SID regardless of number length. Therefore, the complexity of CSID drop to $\mathcal O(1)$ from $\mathcal O(n)$, by applying RevOrder.

The cost of RevOrder for addition and subtraction is quite cheap during both training and inference. In training, RevOrder simply reverses the result digit orders. During inference, almost no additional tokens are required since the recovery of the result sequence can be done with rules.

\subsection{Multiplication and Division}
More complex operations like multiplication and division, can be decomposed to basic operations.

\subsubsection{Multiplication}
Firstly, consider the simplest form of multiplication, nD by 1D, e.g, 12*7=r|48, which consistently requires only 1 SID. This efficiency originates from the definition that 1D by 1D multiplication does not incur any SIDs, with the only one SID being the carry-over number in the addition.

Next, let's examine a more general multiplication example.
\begin{align}
&12 \times 4567 \nonumber \\
=& 12 \times 4000 + 12 \times 500 + 12 \times 60 + 12 \times 7 \\
=& r|00084 + r|0006 + r|027 + r|48 \\
=& (r|00084 + r|0006) + (r|027 + r|48) \\
=& r|00045 + r|408 \\
=& r|40845 \nonumber \\
=& 54804 \nonumber
\end{align}
First, decompose the multiplication as shown in Eqn. (1), which does not require any SIDs (require only count and copy operations that does not use SID in our definition). Second, output the results of each sub-multiplication in reverse order, as demonstrated in Eqn. (2). The zeros in these results can be efficiently generated through a copy operation from previous sequences. The nD by 1D multiplication in reverse order has a CSID of 1. Finally, iteratively combine the adjacent addition results until the final outcome is achieved, as illustrated in Eqn. (3) and (4). As each addition operation involves only two numbers, the CSID remains constant at 1 throughout the process.

In conclusion, the CSID in this multiplication process never exceeds 1, with a complexity of $\mathcal O(1)$.

\subsubsection{Division}
Consider the division \( 948 \div 12 = 79 \):
\begin{align}
&948 \div 12 \nonumber \\
=& 7\ \text{Rem}\ (948 - 12 \times 70) \\
=& 7\ \text{Rem}\ (948 - r|048) \nonumber\\
=& 7\ \text{Rem}\ r|801 \nonumber\\
=& 79\ \text{Rem}\ (r|801-12*9)  \\
=& 79\ \text{Rem}\ (r|801 - r|801)\nonumber\\
=& 79\ \text{Rem}\ (0)\nonumber \\
=& 79\ \nonumber
\end{align}

Utilizing traditional long division alongside RevOrder, the CSID typically remains at 1, with the exception of quotient estimation, as noted in Eqn. (5) and Eqn. (6). Since the CSID analysis here is similar to that of multiplication, we omit it for brevity, . However, it's important to note that quotient estimation often involves heuristic guesswork, making precise CSID measurement challenging. In practice, we observed instances where the language model incorrectly estimated the quotient. To address this challenge, we implemented a rollback mechanism. If an incorrect quotient is detected, as illustrated in Eqn. (7), we insert a symbol 'W' after the line. This serves as a signal to adjust the process and re-estimate the quotient, as demonstrated in Eqn. (8). This method ensures more accurate quotient estimations in the long division process. In practice, a proportion of rollback scenarios are included in training to enhance the model's capability to correct such errors.
\begin{align}
&948 \div 12 \nonumber \\
=& 8\ \text{Rem}\ (948 - 12 \times 80) \nonumber\\
=& 8\ \text{Rem}\ (948 - r|069) \nonumber\\
=& 8\ \text{Rem}\ (-r|21) W\\
=& 7\ \text{Rem}\ (948 - 12 \times 70)  \\
...\nonumber
\end{align}

However, the quotient estimation in division is inherently unpredictable, rendering the CSID of this operation less controllable. Consequently, unlike other arithmetic operations, the CSID for division cannot be consistently maintained at $\mathcal O(1)$. This limitation makes division with RevOrder less robust compared to addition, subtraction, and multiplication, as will be evidenced in our experimental results.

\subsection{Towards More Compact Forms}
To reduce token usage, we propose compact forms while maintaining CSID unchangeable.

For the multiplication example, it can be succinctly rewritten as:
'12\(\times\)4567 = \(12\times\)4000 + \(12\times\)500 + \(12\times\)60+ \(12\times\)7=r|00084 + r|0006 + r|027 + r|48 =   r|00045 + r|408 = r|40845 = 54804'.

Similarly, the division example can be condensed to:
'948\(\div\)12 = 7R - (12\(\times\)70)(r|048)(r|801) \# 9R - (12\(\times\)9)(r|801)(0) = 79',
where R denotes REM and \# denotes a new quotient estimation.

Two principles guide these simplifications:
1. Maintaining CSID: No digits essential for generating subsequent tokens are removed, ensuring the CSID remains unchanged.
2. Eliminating Redundancy: Duplicated digits are removed, but care is taken to avoid introducing ambiguities that might confuse the LM.

\section{Experiments on Arithmetic Problems}
In this section, we aim to address two key research questions (RQs):
\begin{itemize}
  \item RQ1: Does RevOrder enable a language model to function as a reliable calculator? (Section 5.2)
  \item RQ2: Is RevOrder a cost-effective solution for practical using? (Section 5.4)
\end{itemize}

\subsection{Setup}
\subsubsection{Dataset}
Our training dataset is synthetically generated using a Python script, with each sample being an equation formatted with RevOrder, e.g., '123+46=r|961'. The dataset comprises positive integers, except in subtraction where negative numbers may result. Each division equation is assigned a probability of 0.5 to be selected for generating a rollback version. This involves intentionally misestimating a quotient step by a number $\pm 1$, followed by a correction through the rollback process to the accurate estimation. The detailed of the training data is shown in Appendix B.

\subsubsection{Training and evaluation protocol}
We train a model named RevOrder-1B, which has 1.1 billion parameters. This model is trained on the TinyLLaMA 1.1B framework~\cite{zhang2024tinyllama}, utilizing their released finetuning script. Specifically, the learning rate is set to 1e-4 for first 2 epochs and 1e-5 for the last epoch. The batch size is 500.

For evaluation, we employ the BIG-bench Arithmetic sub-task~\cite{srivastava2022beyond} and additional challenging tasks proposed in the GOAT-7B paper~\cite{liu2023goat}. Each task has 1000 equations. We meticulously ensure that there is no overlap between the evaluation datasets and our training dataset, except for unavoidable overlaps in small digits tasks. The evaluation metric is exact match precision.

\subsubsection{Baselines}
As baselines, we compare against three methods:
\begin{itemize}
  \item GOAT-7B~\cite{liu2023goat}: This model, finetuned with 1 million instruction data on LLAMA-7B~\cite{touvron2023llama}, decomposes multiplication and division similarly to our approach. However, it relies on direct result generation for subtraction and addition.
  \item MathGLM-2B~\cite{yang2023gpt}: Finetuned on the GLM-2B model for various arithmetic tasks, MATHGLM-2B claims that extensive step-by-step training data (1m-50m instances) enables GPT models to solve math problems without external calculators.
  \item GPT-4~\cite{openai2023gpt4}: Currently one of the most powerful LMs, GPT-4's results are based on direct mathematical problem-solving, without auxiliary tools or equation decomposition.
\end{itemize}

\subsection{Main Results (RQ1)}

\begin{table*}
    \footnotesize
  \centering
  \begin{tabular}{c|ccccc|ccc}
    \toprule
    Task & \multicolumn{5}{c}{BIG-bench} & \multicolumn{3}{c}{Extra Tasks} \\
    \toprule
     ADD    & 1D    & 2D    & 3D    & 4D    & 5D    & 8D+8D & 16D+8D & 16D+16D \\
     \midrule
    GPT-4 & 100 & 100 & 99.6 & 98.8 & 94.1 & 92.1 & 9.4 & 94.1 \\
    GOAT-7B & 100 & 100 & 99.4 & 98.3 & 98.1 & 97.8 & 97.1 & 97.6 \\
    MathGLM-2B & 100 & 100 & 100 & 100 & 99.4 & - & - & - \\
    RevOrder-1B & 100 & 100 & 100 & 100 & 100 & 100 & 100 & 100 \\
    \midrule
    SUB    & 1D    & 2D    & 3D    & 4D    & 5D    & 8D-8D & 16D-8D & 16D-16D \\
     \midrule
    GPT-4 & 100 & 100 & 99.2 & 98.9 & 92.4 & 70.5 & 10.6 & 59.6 \\
    GOAT-7B & 100 & 100 & 99.7 & 98.6 & 98.4 & 96.8 & 95.8 & 96.3 \\
    MathGLM-2B & 100 & 100 & 99.9 & 99.8 & 98.9 & - & - & - \\
    RevOrder-1B & 100 & 100 & 100 & 100 & 100 & 100 & 100 & 100 \\
    \midrule

    MUL    & 1D    & 2D    & 3D    & 4D    & 5D    & 16D $\times$ 1D & 8D $\times$ 4D & 6D$\times$6D \\
     \midrule
    GPT-4 & 100 & 99.4 & 30.3 & 5.3 & 0.0 & 61.5 & 0.0 & 0.0 \\
    GOAT-7B & 100 & 100 & 97.8 & 96.9 & 96.7 & 99.7 & 88.1 & 96.8 \\
    MathGLM-2B & 100 & 99.9 & 98.3 & 94.9 & 89.9 & - & - & - \\
    RevOrder-1B & 100 & 100 & 100 & 100 & 100 & 100 & 100 & 100 \\
    \midrule

    DIV    & 1D    & 2D    & 3D    & 4D    & 5D    & 16D$\div$1D & 6D$\div$3D & 12D$\div$6D \\
     \midrule
    GPT-4 & 100 & 100 & 94.5 & 90.9 & 53.4 & 54 & 6.4 & 0.0 \\
    GOAT-7B & 100 & 100 & 99.5 & 99 & 96.5 & 99 & 94.1 & 89.3 \\
    MathGLM-2B & 100 & 100 & 99.4 & 100 & 94.9 & - & - & - \\
    RevOrder-1B & 100 & 100 & 100 & 100 & 100 & 99.2 & 100 & 99.4 \\
    \bottomrule
  \end{tabular}
  \caption{Performance comparison on various arithmetic tasks. The results of the baseline methods are taken from their original paper, while the result of GPT-4 is taken from ~\citet{liu2023goat}.}
  \label{tab:performance_comparison}
\end{table*}

The results, as presented in Table 1, demonstrate several key findings. Firstly, RevOrder-1B proves to be a reliable method for addition, subtraction, multiplication, and low-digit division tasks, achieving 100\% accuracy across all corresponding tasks. In contrast, the accuracy of all baseline methods decreases with the increase in digit size. Secondly, while RevOrder-1B shows slight imperfections in large-digit division tasks, it still significantly outperforms baseline models. For instance, RevOrder-1B attains a 99.4\% accuracy on the challenging 12D $\div$ 6D tasks, with an increasing of 10.1\% than that of the best-performing baseline, GOAT-7B.

The major success of RevOrder in multiplication and division can be attributed to its precise execution of basic operations, including addition, subtraction, and nD-1D multiplication. While GOAT-7B and MathGLM-2B also decompose these operations into basic ones, minor errors in these fundamental steps are amplified in subsequent composite operations, leading to a rapid decline in accuracy with larger digits.

In summary, RevOrder emerges as an effective technique, enabling language models to perform exact arithmetic calculations in addition, subtraction, multiplication, and low-digit division tasks.

\subsection{In-Depth Analysis on Division}

\begin{figure}
    \centering
    \includegraphics[width=0.35\textwidth]{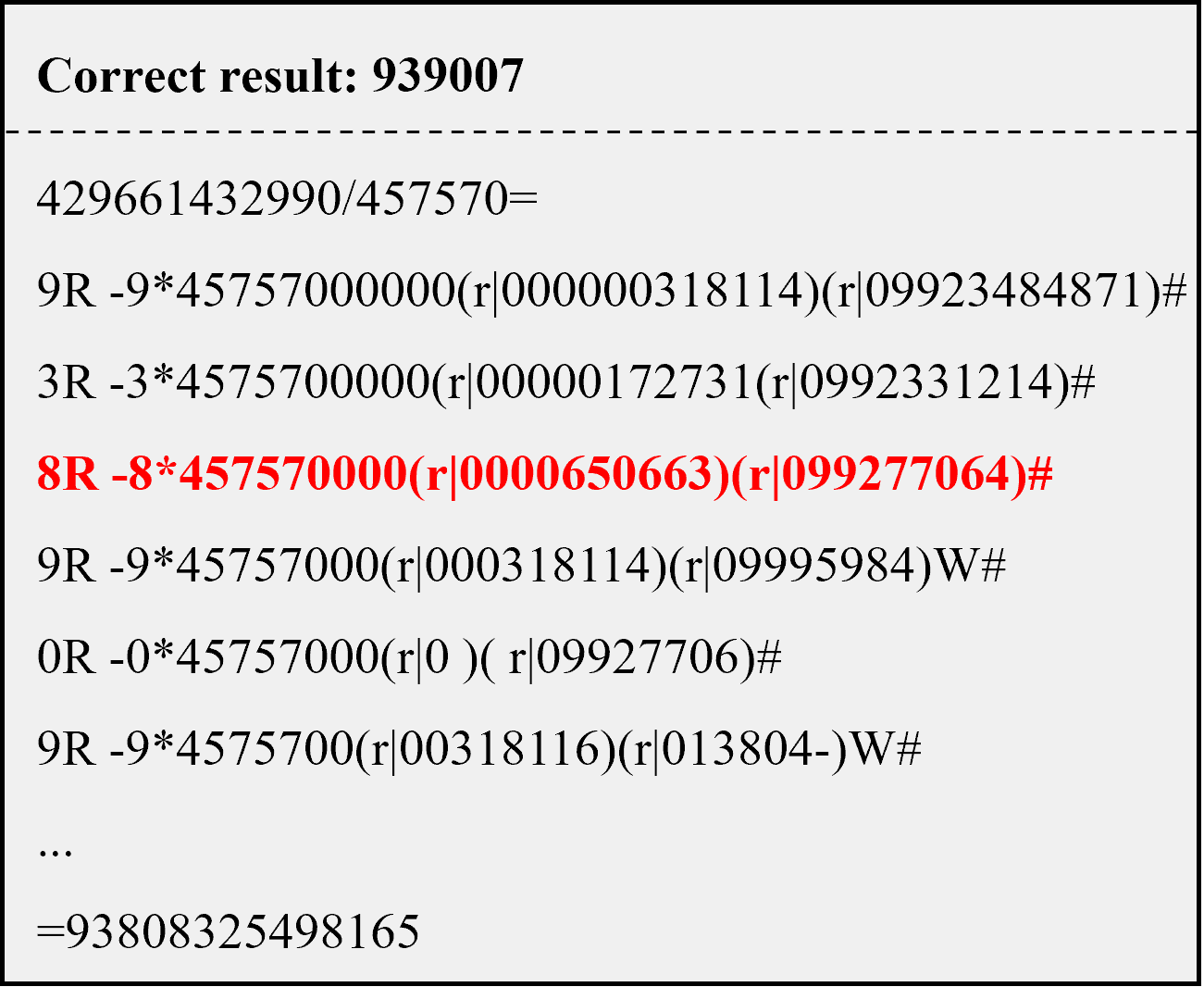}  % Adjust the width as needed
    \caption{An error example of division by RevOrder.}
    \label{fig:my_label}
\end{figure}
Large-digit division represents the sole operation where RevOrder encounters notable difficulties, warranting additional focus.

Upon examining division errors case by case, we discovered that all errors stemmed from incorrect quotient estimations. Fig. 3 illustrates such an error, where RevOrder-1B erroneously estimated the 3rd quotient as 8 (marked in red) instead of 9, without triggering the 'W' symbol for a rollback. Consequently, this led to a series of nonsensical outputs. It's notable that when a constant CSID of 1 is maintained in all four arithmetic operations, no errors occur. Errors only arise during quotient estimation, where CSID is unmeasurable. These results validate  our theory regarding CSID.

\begin{figure}
    \centering
    \includegraphics[width=0.5\textwidth]{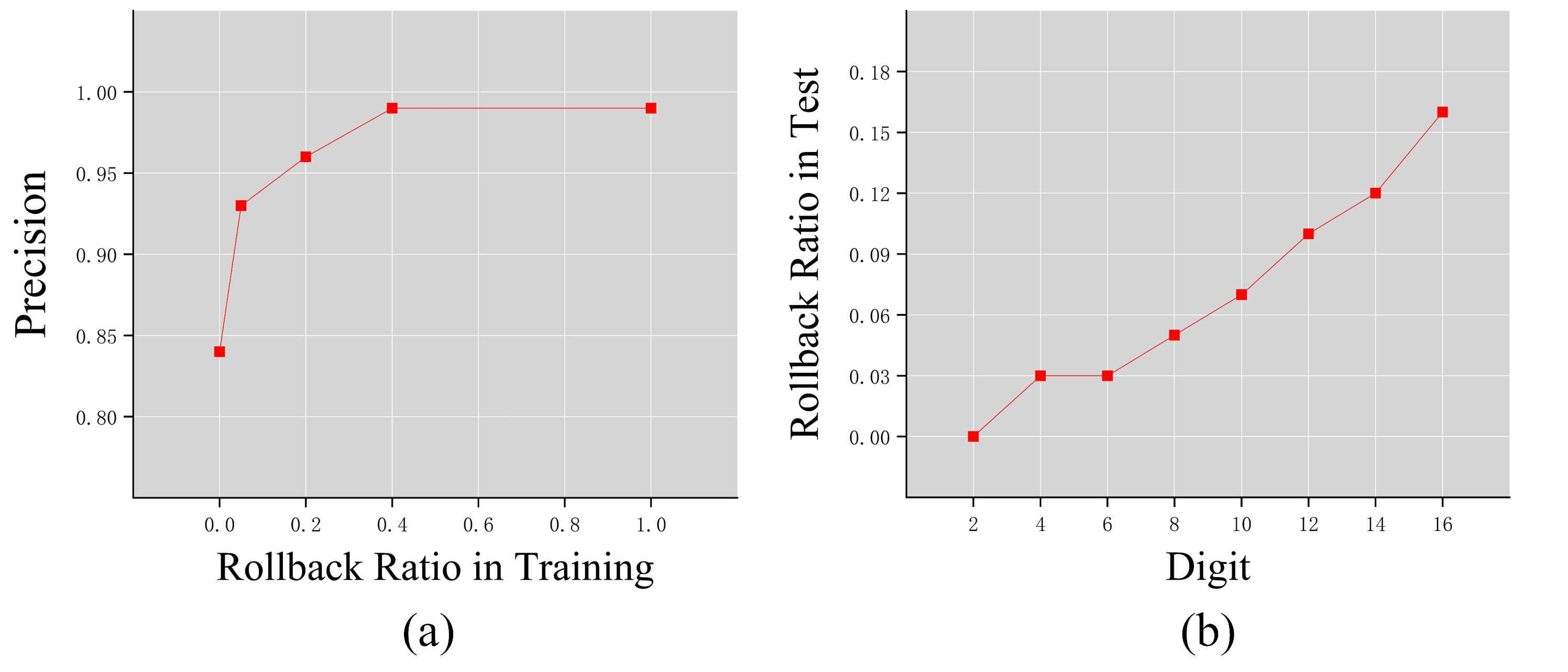}  % Adjust the width as needed
    \caption{Analysis of the rollback ratio in division. (a) Test precision vs. rollback ratio for $12D \div 6D$ division. (b) Probability of rollbacks during testing across different digit sizes. }
    \label{fig:my_label}
\end{figure}

We also assessed the effectiveness of the rollback mechanism. Fig. 4(a) presents the test precision for $12D \div 6D$ division across varying rollback ratios. A stark precision decline to 0.84 is observed with no rollback (ratio = 0). Precision does not significantly improve when the ratio exceeds 0.4, though this is partly due to the high baseline precision of 0.99. Fig. 4(b) illustrates the frequency of rollbacks during testing, indicating a higher incidence of rollbacks with larger digits. This trend underscores the importance of the rollback technique, particularly as it compensates for the increased likelihood of errors in quotient estimation with larger numbers.

\subsection{The Cost of RevOrder (RQ2)}
\subsubsection{Cost of Training}
By maintaining a low CSID, RevOrder simplifies the learning process for arithmetic problems, thereby reducing the volume of training data required. Table 2 compares the number of training equations needed for various methods. Despite being a smaller model, RevOrder-1B achieves perfect precision with at most half the training equations compared to other methods. Recent studies indicate that larger models often require less training data for task mastery~\cite{hoffmann2022training,xia2022training}. Consequently, the training cost advantage of RevOrder is likely to be even more pronounced with larger LLMs.

\begin{table}
  \centering
  \begin{tabular}{ccc}
    \toprule
    Model & \# Equations & 100\% ACC  \\
    \midrule
    RevOrder-1B & 0.5m  & Yes\\
    MathGLM-2B & 1m-50m & No\\
    GOAT-7B & 1.7m & No\\

    \bottomrule
  \end{tabular}
  \caption{Number of training equations for different methods. This table reports the dataset size required for RevOrder-1B to achieve 100\% accuracy on all Big-bench arithmetic sub-tasks. \# Equations denotes the number of training equations.}
  \label{tab:performance_comparison}
\end{table}

\subsubsection{Cost of Inference}
The inference cost is assessed based on the number of additional tokens required for performing arithmetic calculations with RevOrder. We make two assumptions: 1) Each character (digit, symbol, etc.) is counted as one token, and 2) if the final result is output in reverse, the recovery process is handled by the tokenizer's decode function.

For addition and subtraction equations, only one extra token ('r|') is required. For multiplication and division equations, the number of extra tokens used is illustrated in Fig. 5. RevOrder is more token-efficient in both types of equations. Firstly, the compact form introduced in Section 3.3 significantly reduces the token requirement for division, approximately halving the number of extra tokens. Secondly, the iterative combination approach in multiplication, as exemplified in Eqn. (3), also notably reduces token usage in multiplication.

\begin{figure}
    \centering
    \includegraphics[width=0.45\textwidth]{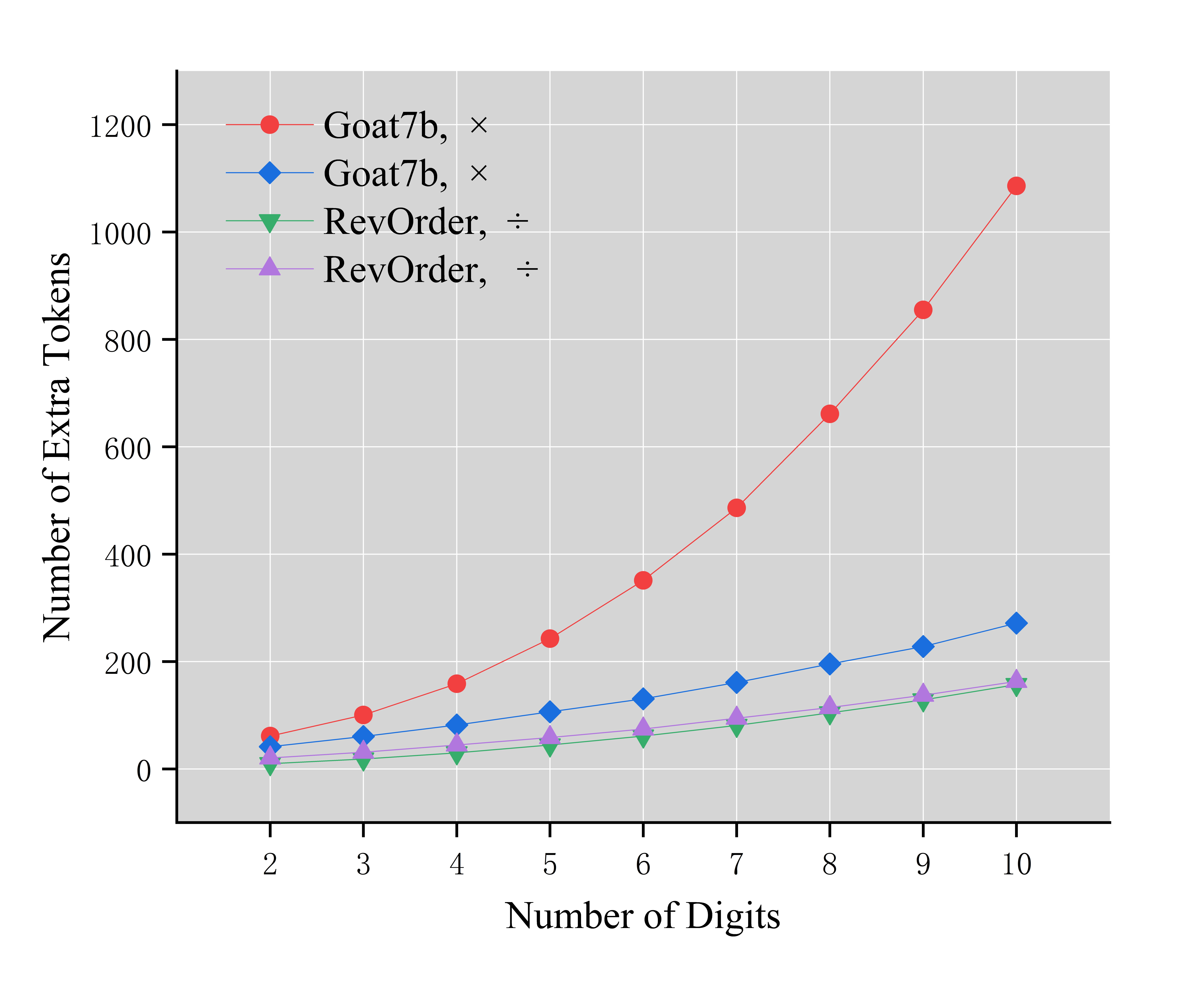}  % Adjust the width as needed
    \caption{The number of extra tokens required for multiplication and division.}
    \label{fig:my_label}
\end{figure}

However, it must be acknowledged that for large-digit multiplication and division tasks, the token consumption of RevOrder increases polynomially and may eventually exceed the cost of using external tools. LLM service providers can set a threshold of digit number to decide between RevOrder and tool-based solutions.

\section{Additional Experiments on Math Word Problems}
In this section, we delve into finetuning scenarios to address the research question:
\begin{itemize}
\item RQ3: How does applying RevOrder affect finetuning performance on mathematical tasks?
\end{itemize}

\subsection{Setup}
The experiment is conducted on GSM8K~\cite{cobbe2021training}, a dataset of 8.5K high quality linguistically diverse grade school math word problems created by human problem writers. Our experiments utilize LLAMA2-7B~\cite{touvron2023llama} as the foundational model. We modified the equations in the GSM8K training set to adopt the RevOrder format. This adaptation involved two major updates: Firstly, we presented the outcomes for addition, subtraction, and multiplication in reverse order. Secondly, polynomial equations were expanded and solved iteratively, in pairs. Noted that we did not decompose multi-digit multiplications and divisions, as these cases are infrequent in the GSM8K dataset.  To further enhance the model's proficiency with RevOrder, we supplemented the training set with a small, synthetically generated dataset using a Python script. The comprehensive details of the dataset and the training parameters are provided in Appendix C.

\subsection{Results}
\begin{table}
  \centering
  \begin{tabular}{ccc}
    \toprule
     & Baseline & RevOrder \\
    \midrule
    Score & 41.6  & 44.4 (+2.8)\\
    \midrule
    Equation Acc & 88.9 & 94.1 (+5.2)\\
    Acc of + & 96.7 & 99.8 (+2.1)\\
    Acc of - & 97.0 & 99.6 (+2.6)\\
    Acc of * & 95.8 & 98.8 (+3)\\

    \bottomrule
  \end{tabular}
  \caption{Fine-tuning results on GSM8K Dataset. This table compares the performance of models fine-tuned with the original GSM8K dataset (baseline) against those finetuned using the RevOrder-modified GSM8K dataset. The Score is measured by the correctness ratio of final results.}
  \label{tab:performance_comparison}
\end{table}

From the analysis, it is evident that RevOrder significantly reduces calculation errors, by 94\% for addition, 87\% for subtraction, and 46\% for overall equation errors, thereby enhancing the final score. This improvement underscores the potential of seamlessly integrating RevOrder into fine-tuning processes to achieve substantial performance gains.

We also observe the errors, and find most of the errors are due to lack of enough training. Therefor, the model cannot well follow the instructions of RevOrder. Some examples are presented in Appendix C.

Hence, integrating RevOrder effectively into LMs is ideally conducted during the pretraining stage rather than the fine-tuning stage. The primary rationale is that excessive fine-tuning can lead to catastrophic forgetting, thereby impairing the general capabilities of LMs~\cite{luo2023empirical,ramasesh2021effect}.

\section{Conclusion}
In this paper, we introduce the CSID as a metric to evaluate the complexity of arithmetic equations and demonstrate that even large-scale LLMs struggle with high-CSID equations. We propose RevOrder, an innovative technique that ensures accurate arithmetic calculations by minimizing CSID, thereby enhancing precision while reducing both training and inference costs. Our experiments confirm that RevOrder significantly outperforms previous methods in terms of accuracy and efficiency.

For future work, we identify two possible paths: Firstly, developing token-efficient decomposition algorithms suitable for larger LLMs, which can handle higher CSIDs for complex arithmetic operations. Secondly, integrating RevOrder into LLMs' pretraining could enhance arithmetic capabilities more fundamentally than finetuning, reducing the risk of catastrophic forgetting and ensuring broader model proficiency.

Ultimately, RevOrder stands out as a particularly promising approach for arithmetic operations, especially addition and subtraction, due to its precision and efficiency. This positions it as a competitive alternative to existing methods in enhancing LLMs' arithmetic reasoning.
\newpage

\bibliography{custom}
\newpage

\appendix
\section{The CSID Analysis of Multiplication and Division}
\label{sec:appendix}
This section extends the CSID analysis to nD by nD multiplication and nD by mD division, following the algorithmic approach outlined in Section 4.2 but excluding the RevOrder technique.

\subsection{Multiplication}
The decomposition of an nD by nD multiplication into n sub-multiplications, each an nD by 1D operation, serves as the initial step. This phase does not generate SIDs, as all required digits for $a \times b$ are immediately accessible.

Addressing these sub-multiplications yields up to $n^2 + n \times (n+1) = 2n^2 + n$ SIDs, with $n^2$ SIDs allocated for the sub-multiplications and $n \times (n+1)$ SIDs dedicated to storing the outcomes.

Aggregating the results of these sub-multiplications necessitates a maximum of $4n^2$ SIDs, with each addition consuming $4n$ SIDs, $2n$ for carry-overs and another $2n$ for storing the results.

Consequently, directly generating an nD by nD multiplication outcome requires a maximum of $6n^2 + n$ SIDs, indicating a complexity of $\mathcal{O}(n^2)$. This substantial complexity explains the difficulty models face with even 2D by 2D multiplications.

Decomposition methods, as applied in models like GOAT-7B and MathGLM-2B, reduce the CSID to $\mathcal{O}(n)$, by omitting intermediate results from the SID count, though carry-overs are still considered.

\subsection{Division}
For an nD by mD division, typically $n-m$ iterations are needed, each estimating a quotient digit.

Each iteration involves an nD by 1D multiplication and a subtraction, with the multiplication incurring 2m SIDs for result and carry-over digit storage, and the subtraction using up to 2n SIDs for result storage and borrow digits.

Thus, the total CSID for an nD by mD division reaches $(2m+2n)*(n-m)=2n^2 - 2m^2$, amounting to a complexity of $\mathcal{O}(n^2-m^2)$.

This estimation excludes the quotient estimation step's complexity, which could further complicate large number divisions, potentially surpassing the $\mathcal{O}(n^2-m^2)$ complexity.

In models like GOAT-7B and MathGLM-2B, using decomposition methods keeps the CSID at $\mathcal{O}(n)$, with the subtraction's borrow digits being the primary complexity factor.

\section{Training Data for Arithmetic Experiments}
\begin{figure}
    \centering
    \includegraphics[width=0.4\textwidth]{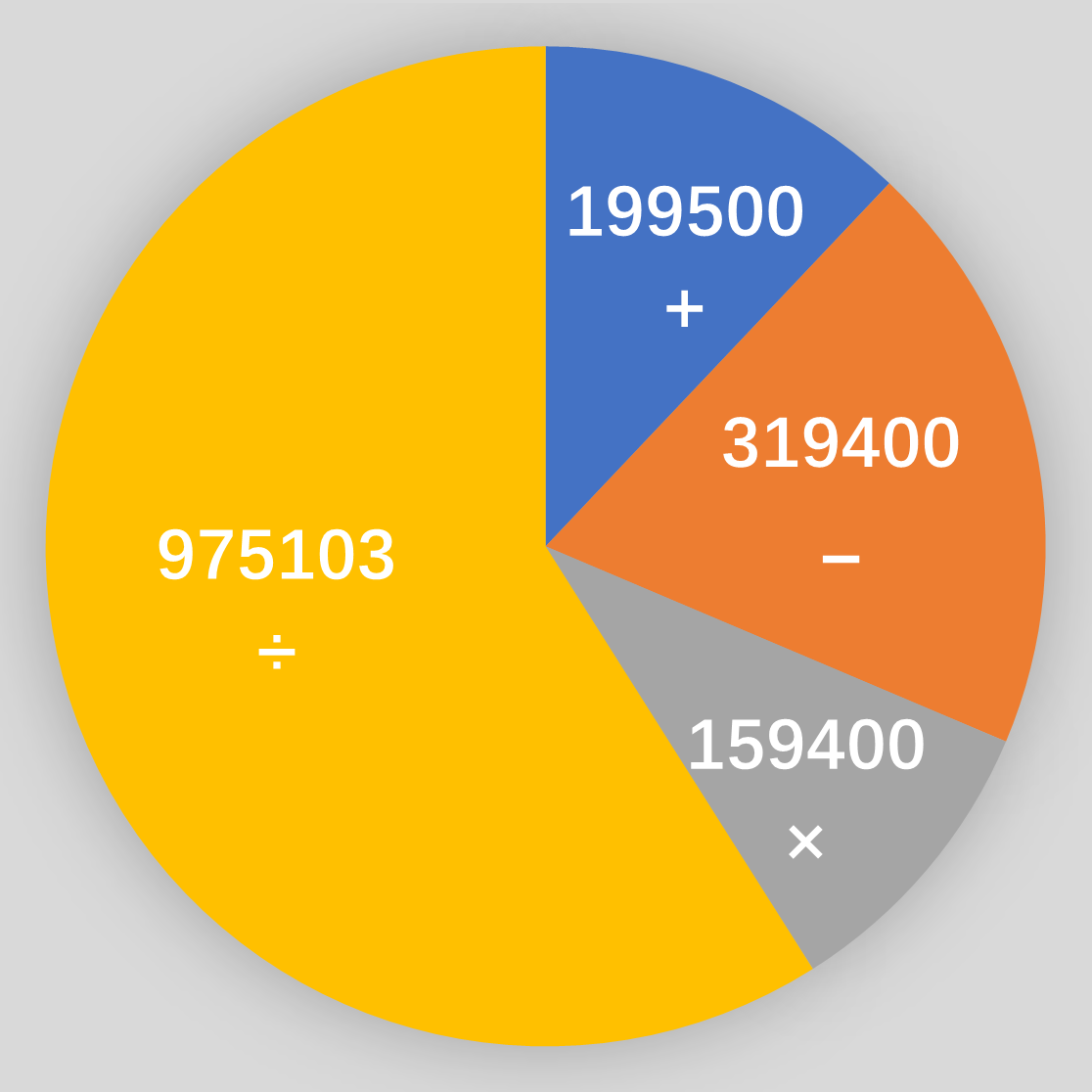}  % Adjust the width as needed
    \caption{The distribution of the equations in training set.}
    \label{fig:my_label}
\end{figure}

The training dataset comprises 1.7 million equations. For addition and subtraction tasks, equations involve numbers as large as 16D on both sides. Multiplication tasks are capped at 8D by 8D, supplemented by 16D by 1D equations to enhance generalization in the test set. Division tasks feature dividends up to 16D. Fig. 6 illustrates the distribution of these equations. The major training samples are division, since the quotient estimation steps require more training samples to achieve a high precision.

\section{Settings for Math Word Experiments}
\subsection{Training Data}
Our approach involved two types of instructional data to train models on arithmetic tasks using RevOrder.

Firstly, we modified the original GSM8K dataset to reflect RevOrder formatting. An example of this adaptation is illustrated in Fig. 7.

\begin{figure*}
\centering
\includegraphics[width=0.9\textwidth]{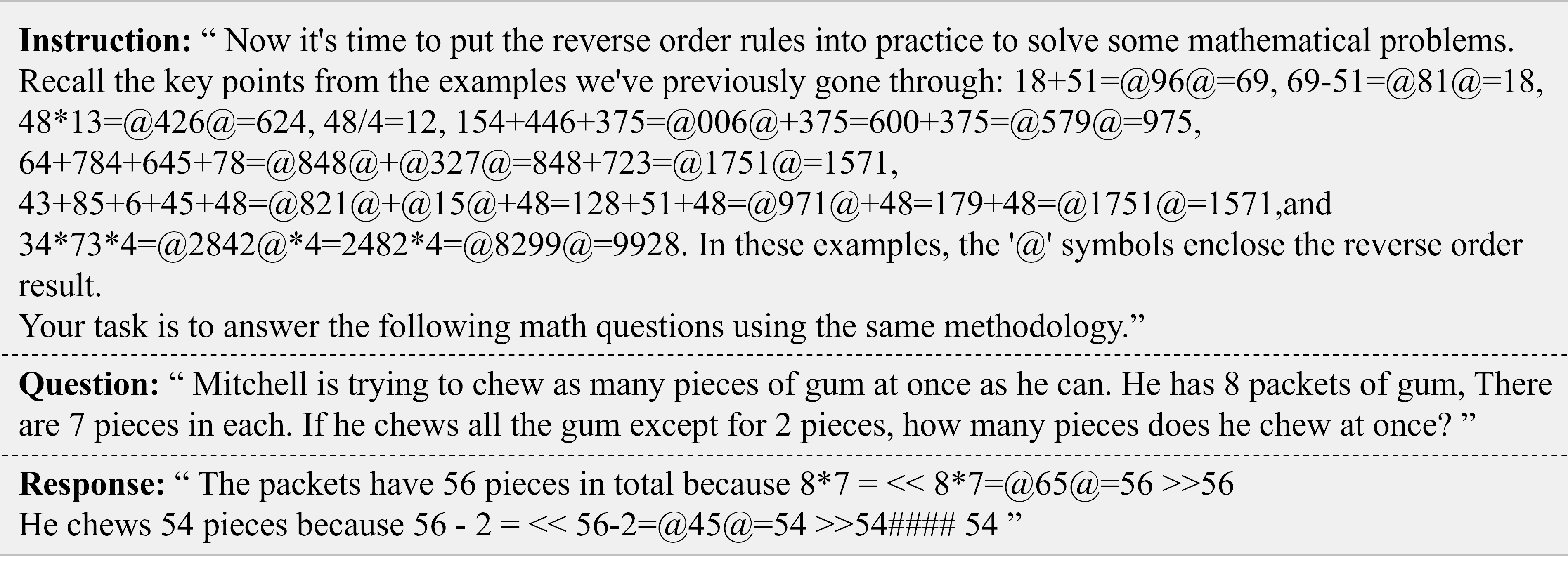}
\caption{A data sample from the GSM8K dataset formatted in RevOrder.}
\label{fig:gsm8k_revorder}
\end{figure*}

Secondly, to further bolster the model's proficiency in RevOrder calculations, we compiled an additional enhancement dataset. A sample from this dataset is depicted in Fig. 8.

\begin{figure*}
\centering
\includegraphics[width=0.9\textwidth]{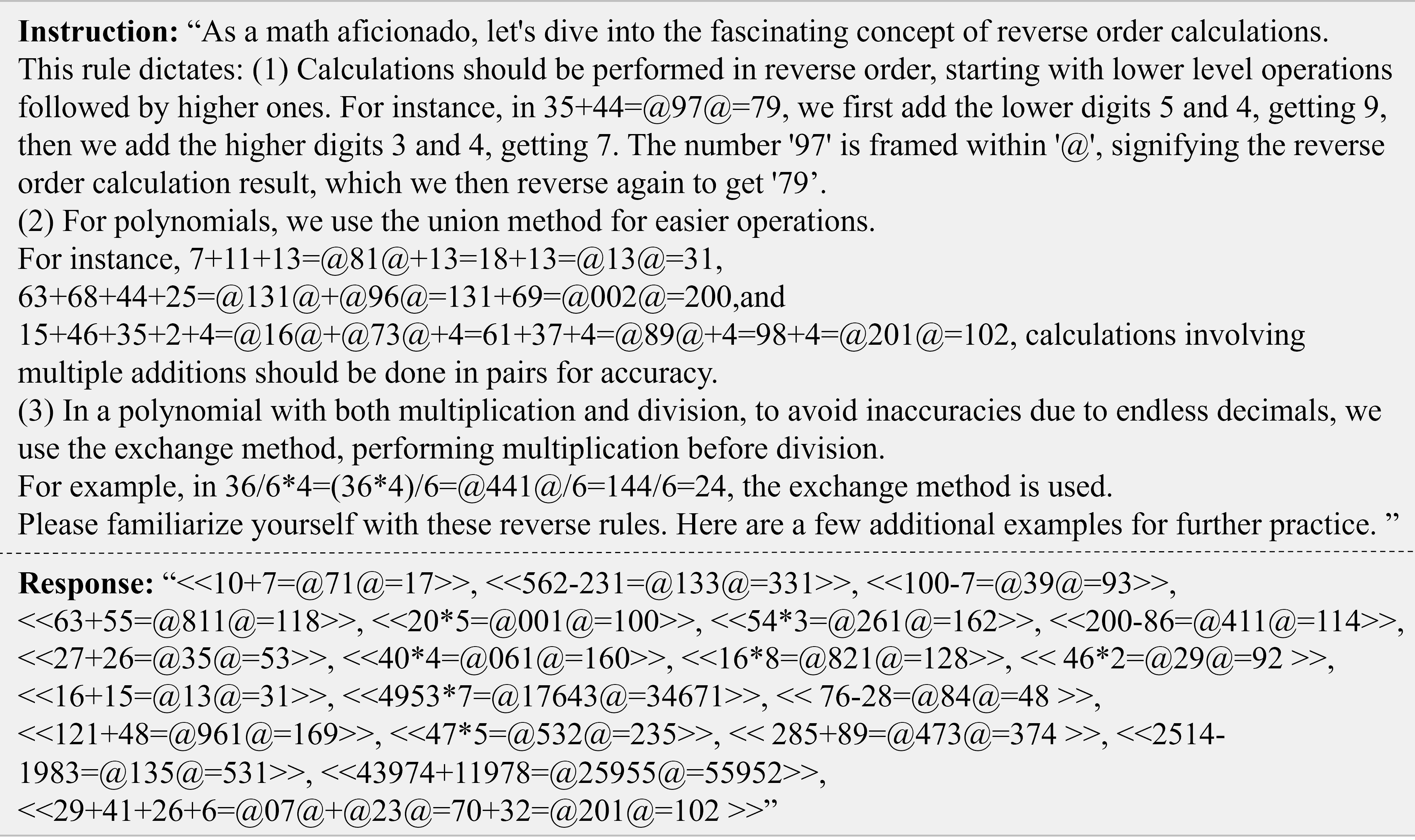}
\caption{A sample from the additional enhancement dataset for RevOrder calculations.}
\label{fig:enhancement_dataset}
\end{figure*}

Given the limited size of the training data, the 7B model faced challenges in mastering the use of the reverse symbol r|. To address this, we introduced a notation where all numbers enclosed by @@, signify reverse order.

\subsection{Training Details}
The models were trained with a batch size of 32 and a learning rate of 5e-5, employing a warm-up ratio of 0.08 over 3 epochs. During each epoch, the model was exposed to both the additional datasets and the GSM8K datasets sequentially.

\subsection{Equation Errors}
Fig. 9 showcases representative errors encountered in the GSM8K test set, attributable to difficulties in adhering to RevOrder instructions. For instance, while the model successfully solved the second equation in reverse order, it faltered in performing the simple task of reversing the solution to arrive at the final result. 

\begin{figure*}
\centering
\includegraphics[width=0.9\textwidth]{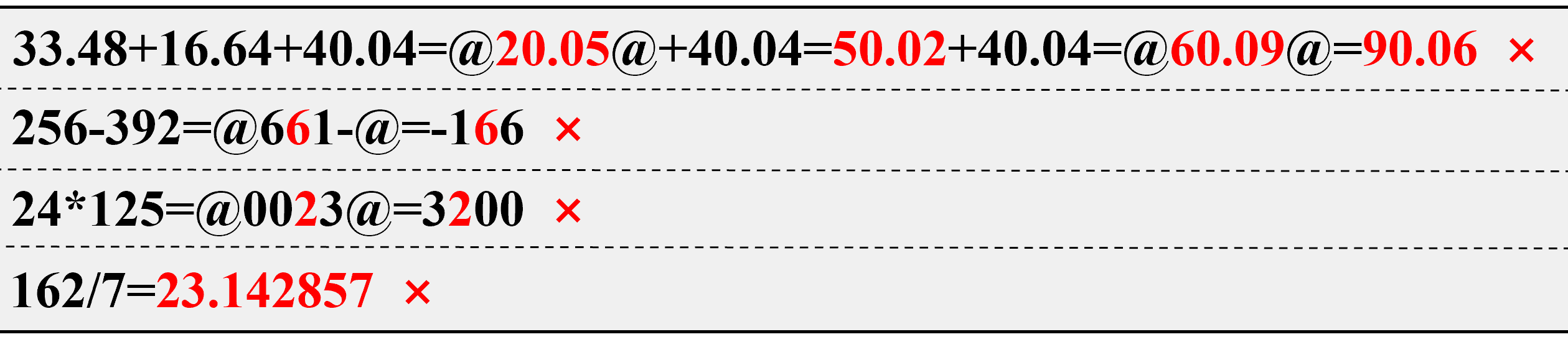}
\caption{Illustrative errors from the GSM8K test set encountered by the model trained with RevOrder.}
\label{fig:errors_gsm8k}
\end{figure*}

\end{document}